**Development of a Reliable and Accessible Caregiving Language Model (CaLM)**

Bambang Parmanto, Bayu Aryoyudanta, Wilbert Soekinto, I Made Agus Setiawan, Yuhan Wang, Haomin Hu, Andi Saptono, Yong K. Choi.

Department of Health Information Management, University of Pittsburgh.

## Abstract

Introduction: One in five adults in the US currently serves as a family caregiver for an individual with a serious illness or disability. Unlike professional caregivers, family caregivers often assume this role without formal preparation or training. Because of this, there is an urgent need to enhance the capacity of family caregivers to provide quality care. Leveraging technology as educational tools or as adjunct to care is a promising approach that has the potential to enhance the learning and caregiving capabilities of family caregivers. Large language models (LLM) can potentially be used as a foundation technology for supporting caregivers. LLMs fall into a category called Foundation Models (FMs), a large-scale model trained on a broad data set that can be adapted to a range of different domain tasks. Despite their potential, FMs have a critical weakness known as "hallucination"" where the model generates information that can be misleading or inaccurate. The reliability of information is essential when language models are deployed as a front-line help for caregivers.

Objective: This study aimed to (1) develop a reliable Caregiving Language Model (CaLM) by using FMs and a caregiving knowledge base, (2) develop an accessible CaLM using a small FM that requires fewer computing resources, and (3) evaluate the performance of the model compared to a large FM.

Methods: We developed CaLM using the Retrieval Augmented Generation (RAG) framework combined with FM fine-tuning for improving the quality of FM answers by grounding the model on a caregiving knowledge base. The key components of CaLM are the Caregiving Knowledge Base, a Fine-tuned Foundation Model (FM), and a Retriever module. We used two small FMs as candidates for the foundation of CaLM (LLaMA-2 and Falcon with 7 billion parameters) and a large FM GPT-3.5 (estimated 175 billion parameters) as a benchmark. We developed the caregiving knowledge base by gathering various types of documents from the Internet. In this study, we focused on caregivers of individuals with Alzheimer's Disease Related Dementias (ADRD). We evaluated the models' performance using the benchmark metrics commonly used in evaluating language models and their reliability to provide accurate references with the answers.

Results: The RAG framework improved the performance of all FMs used in this study across all measures. As expected, the large FM performed better than small FMs across all metrics. The most interesting result is that small fine-tuned FMs with RAG performed significantly better than GPT 3.5 across all metrics. The fine-tuned LLaMA-2 small FM performed better than GPT 3.5 (even with RAG) in returning references with the answers.

Conclusion: The study shows that reliable and accessible CaLM can be developed by using small FMs with a knowledge base specific to the caregiving domain.

## Introduction

The number of family caregivers for people with complex and chronic conditions, such as dementia and other disabilities, is increasing dramatically. In 2020, an estimated 53 million adults in the United States (one in five) served as family caregivers, up from the estimated 43.5 million in 2015 [1]. These numbers are expected to climb dramatically over the next 30 years [1, 2]. Unlike professional healthcare providers and caregivers, family caregivers often assume this role without formal education or training [3], leaving them underprepared for the complex tasks of caregiving. This lack of preparation can lead to increased stress and a sense of being overwhelmed [4]. Previous studies show that family caregivers are at risk for poor psychological health, physical health, and quality of life; strains in family relationships; and restrictions in social and work participation [5-8]. There is an urgent need to enhance the capacity of caregivers, although there remain unanswered questions of how best to support the diverse needs of family caregivers [9]. One of the key features of successful interventions in supporting caregivers is to equip the caregivers with practical knowledge and skills for providing care, including knowledge about the care recipient's condition, associated symptoms and their progression, as well as skills that enable them to address the needs of the care recipients [10]. In this context, technology can play a pivotal role in supporting caregivers as a means of delivering educational tools or serving as a supplementary aid in the caregiving process [2, 11-14]. Notably, two of the ten research priorities identified by the Summit on Family Caregiving focus on utilization of technology for family caregiving, although technology can also be used to support the remaining eight research priorities [2].

Recent advances in generative artificial intelligence (GAI) [15-19] have resulted in the popularity and widespread use of large language models (LLM), such as ChatGPT, captivating global interest with their ability to generate intelligent and context-aware responses to a wide spectrum of users' questions or prompts [18, 20-22]. These LLMs fall into a category called Foundation Models (FMs) [23]. A FM is a large-scale machine-learning model trained on diverse and comprehensive data sets. These data sets equip the FMs with versatility to perform a wide range of tasks across different domains [17, 19, 23]. FM provides a base, or a foundation, on which other specific models can be built. FMs such as OpenAI's GPT-3 and GPT-4 are pre-trained using diverse corpora of contents found across the Internet. These pre-trained models can serve as the basis for developing educational content as well as interactive agents, such as chatbots, to support caregivers [24]. The interactive agents will be able to address common requests from caregivers, including answering questions about the care recipient's condition, associated symptoms, and symptom progression, as well as teaching skills that enable caregivers to address the needs of the care recipients [25].

However, FMs often fail to answer domain-specific questions that require factual knowledge. The responses generated by these models, while impressive and convincing, can be misleading or completely wrong—a phenomenon called hallucination [26]. This issue is particularly problematic because it may be inherent to LLM even when the size gets larger: it is a feature, not a bug [26, 27]. This means that the system cannot be fully trusted in contexts where accuracy is paramount, such as in caregiving. In such contexts, the reliability and factual accuracy of information are non-negotiable, as they underpin decisions that have direct consequences on the health and well-being of individuals. Furthermore, aside from hallucination, even the most powerful pretrained FM will most likely not meet the specific needs of caregivers right out of the box.

Adaptations have been developed to equip FMs to meet the specific needs of particular tasks, and three of the most prominent adaptation methods are: prompt engineering, fine tuning, and most recently Retrieval Augmented Generation (RAG) [28-30]. Prompt engineering is the most popular method, and its goal is to guide the model towards desirable answers. This is the simplest approach because it does not involve retraining the FM or developing a knowledge base. RAG, on the other hand, introduces additional layers that use external knowledge (data sources) to provide the context for improving the performance and relevance of FM. It is more complex to implement than prompt engineering because it requires the development of domain knowledge. Fine-tuning is the most complex method in terms of implementation because it requires retraining the FM. By strategically combining these methods, it is possible to enhance an FM's functionality, tailoring it to deliver more precise and useful outputs for domain-specific applications such as caregiving.

### Objectives

The objective of this exploratory study is to develop a reliable and accessible Caregiving Language Model (CaLM). To achieve reliability, CaLM will use RAG framework that employs a caregiving knowledge base to generate prompts to provide a caregiving context to any questions from users. CaLM is further fine-tuned by retraining the FM with caregiving-related data that trains the FM to provide authoritative references and informed answers to caregiving related questions. To achieve the goal of accessibility, CaLM uses a small FM that can be deployed with a modest computing infrastructure in a home or a small organization. CaLM can further be used to develop downstream technologies, such as a caregiving chatbot, aiming to support caregivers in various settings.

## Methods

### Caregiving Language Model (CaLM) Architecture

The overall architecture of the Caregiving Language Model (CaLM) is illustrated in Figure 1. The key components of CaLM are the Caregiving Knowledge Base, a Fine-tuned Foundation Model (FM), and a Retriever module. The knowledge base and retriever module are part of the Retrieval-Augmented Generation (RAG) framework [28]. An interaction system can be added if CaLM is implemented as a conversation agent such as chatbot. CaLM uses a RAG framework to give the FM access to information that is specific to caregiving, and it uses fine-tuning to further retrain the FM to answer questions related to caregiving. Each of the modules will be discussed further in this method section.

The most common interaction between caregivers and language models is an open-ended questions-and-answers (Q-A) method [28]. In a regular FM (which we call Vanilla FM), a question from caregivers will be submitted to an FM, and the FM will retrieve answers based on the pretrained knowledge representation of the FM. In CaLM, a question from a user is appended by a prompt generated from the Caregiving Knowledge Base before it is submitted to the FM. The Retriever module in CaLM retrieves semantically similar information from the Caregiving Knowledge Base and creates prompts to accompany the user question. The FM uses the question and the prompts to get a more relevant answer than it could without the prompts.

In CaLM, the FM is further fine-tuned by retraining the FM using a Q-A training set containing common questions and answers related to caregiving. Fine tuning can be used to impart FMs with domain-specific terminology and with nuance or personalization to a specific population. This technique can be used to train FMs in areas important to caregiving, such as empathy. The question from the caregiver, combined with prompts from RAG, is submitted to the fine-tuned FM, which subsequently provides answers that are more accurate and more knowledgeable about caregiving.

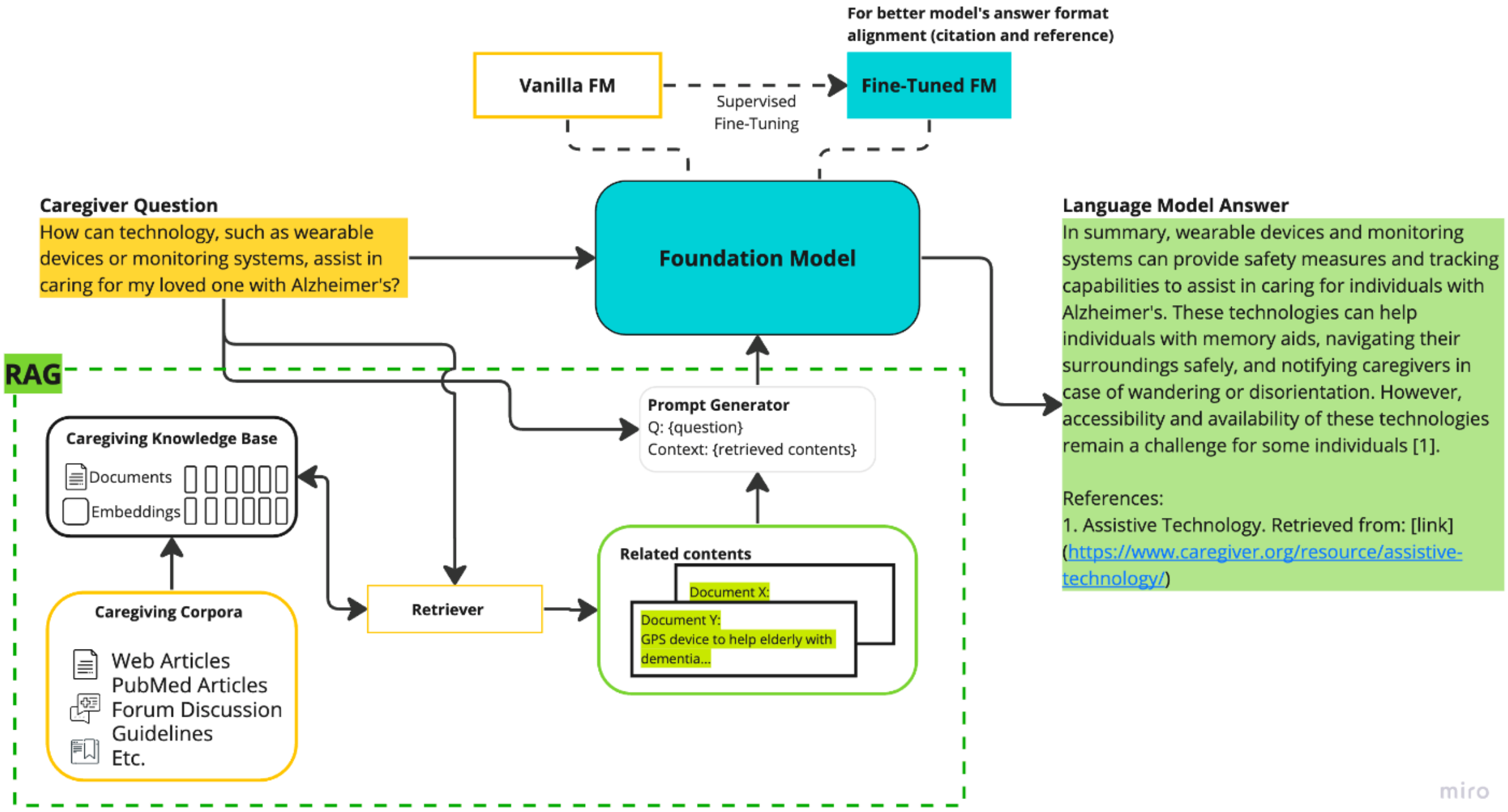

*Figure 1. Caregiving Language Model (CaLM) Architecture*

## Foundation model

Large Language Models (LLMs), such as ChatGPT, are a subset of an artificial intelligence system called a Foundation Model (FM). FMs are designed to be general purpose models capable of performing many different tasks and can also be adapted to a variety of tasks. As the name implies, we will use FMs as the base foundation for CaLM. Hundreds of FMs are currently available, and a recent paper catalogs more than 75 major transformer-based models alone [31]. The goal of this project is to build a reliable and accessible model in the caregiving domain; therefore, the FM needs to fulfill two requirements: good performance and relatively small size. In this project, we evaluate three different FMs summarized in Table 1. Falcon and LLaMA 2 were chosen because the models were at the top Open LLM Leaderboard [32] when this study was carried out. LLaMA (Large Language Model Meta AI) [33] is a family of LLMs released by Meta starting in February 2023. Falcon is a family of LLMs developed by the Technology Innovation Institute (TII) based in Abu Dhabi, United Arab Emirates [34].

We chose FMs that are considered "small" LLMs with 7 billion parameters. If CaLM can be developed using a "small" FM, then it will make powerful AI capabilities more accessible, affordable, and versatile. The models can be deployed using regular computing infrastructure available in low-resource settings such as small community organizations or homes. The open source FMs used in this project have permissive licensing models that can potentially be deployed in low-resource settings. We also included a "large" LLM, a proprietary GPT-3.5 with an estimated 175 billion parameters owned by OpenAI [17, 35, 36], as a point of comparison to showcase the potential small FMs have for developing CaLM.

Table 1. Foundation Models (FMs) used in the study

| Foundation Model | Developer | Description | Licensing | Parameters |
|---|---|---|---|---|
| LLaMA-2 7B | Meta | Large Language Model Meta AI. Top-ranked model | Open source | 7 billion |
| Falcon 7B | Technology Innovation Institute, UAE | Part of Falcon LLM Family, Apache permissive license. Top-ranked model | Open source | 7 billion |
| GPT 3.5 | OpenAI | Generative Pre-trained Transformer | Proprietary | Est. 175 billion |

## Caregiving Corpus and Knowledge Base

The Knowledge Base is central to CaLM in providing and guiding the FM with a rich context of caregiving-specific information. The long-term goal of this study is to develop CaLM for various care recipients' conditions. In this preliminary work, we focused on developing CaLM for caregivers of individuals with Alzheimer's disease and related dementias (ADRD). The development of the Caregiving Knowledge Base started with the development of the caregiving corpora. In this project, we developed a caregiving corpus related to ADRD by gathering a significant collection of data related to ADRD caregiving. This data included publicly available journal articles, care guidelines, and practical insights from online caregiver discussion forums. Table 2 describes the data sources that were collected for the knowledge base.

ADRD were selected for the proof-of-concept development of the CaLM due to the substantial and growing impact these conditions have on the global population, especially within the aging demographic [37, 38]. The intricate care requirements associated with the cognitive and behavioral symptoms of ADRD present a complex challenge that caregivers must navigate, often without formal training [7, 39]. Given the progressive nature of these conditions, caregivers are in need of long-term support and strategies, underlining the importance of a dedicated resource like CaLM. By addressing ADRD, the model can provide substantial support to a vast community of caregivers who are frequently underserved when it comes to specialized care resources. Moreover, with an extensive amount of research and guidelines already available for ADRD, there is a rich foundation upon which to build a detailed and accurate knowledge base, making it a particularly suitable focus for the initial deployment of this innovative tool.

The documents we collected were crawled and downloaded from the websites in various formats such as HTML, PDF, or plain text (Table 2). These raw documents then underwent several pre-processing stages before they were converted into a caregiving corpus. The pre-processing procedure included data format conversion, text cleaning, and document chunking. During data format conversion, documents in HTML and PDF files were converted into either plain text markdown format or plain text. The converted documents were cleansed from errors of duplication and unnecessary characters such as extra spaces, new lines, punctuations, and non-ASCII tags during conversion using regular expression (Regex) rules.

The last stage of text processing was document chunking and converting chunk documents into Document format with metadata. The document chunking broke down the documents in the caregiving corpus into smaller “chunks” with a length of 1200 characters each to fit the FM context window limitation and reduce unrelated text in the generated prompt in the RAG system. These chunks of text were then converted into a 768 dense dimensional vector using Siamese BERT-Network [40] as the embedding model. We utilized BGE, a recently released embedding model that was pretrained on massive datasets and multiple tasks as a general-purpose embedding model [41]. BGE was selected because it provides a good trade-off on the memory usage, speed of computation, and embedding quality. These vectors were then stored in a Chroma DB vector database and function as the retriever database for the RAG system. Chroma DB was selected in this experiment for the ease of its setup; it is not too strict on the data schematics and supports multiple built-in distance functions such as Squared L2, Inner Product, and Cosine similarity.

Table 2. Sources for Caregiving Corpus

| **Source** | **Source File Format** | **Type** | **Document extracted** | **Number of Documents** | **Number of Chunks** |
|---|---|---|---|---|---|
| caregiver.com | HTML | Online caregiving forum, discussion, and tips | High-quality questions-answers | 142 | 1,591 |
| agingcare.com | HTML | Caregiver support website | Caregiving resources and questions-answers | 402 | 4,169 |
| alzconnected.org | HTML | Discussion forums from Alzheimer’s Association | Low-Technical Resources in Question Answer Format | 4,087 | 4,087 |
| deliriumnetwork.org | PDF | Repository of resources related to delirium | High-quality literatures and resources | 1,714 | 72,426 |
| PubMed | PDF | Database of journal articles | High-quality literatures and resources | 2,195 | 114,383 |
| nia.nih.gov | PDF + HTML | Education resources on aging and ADRD | Literatures and caregiving resources | 9 | 133 |
| alzheimers.gov | HTML | Education resources on aging and ADRD | Literatures and caregiving resources | 8 | 55 |
| alzheimers.org.uk | HTML | Education resources on aging and ADRD | Literatures and caregiving resources | 7 | 39 |
| alz.org | HTML | Education resources on aging and ADRD | Literatures and caregiving resources | 2 | 12 |
| Other web sources | HTML | web article | Caregiving Resources | 2 | 31 |
| Total | | | | **8,568** | **196,926** |

## Retriever Module in CaLM: Providing Caregiving Context Prompts

The Retriever module in CaLM searches for semantically related information from the Caregiving Knowledge Base to provide a caregiving context that matches users' questions. The related caregiving information provides enhanced context that is appended to the user's prompt and passed to the FM. It is important to implement a retrieval mechanism that can efficiently search through the Caregiving Knowledge Base to respond to a user's questions. A Dense Passage Retriever (DPR) [42] is used to develop the retriever module in CaLM. DPR captures more complex semantic relationship between the query and the documents, leading to more accurate retrieval results.

In the retriever implementation, the user's question is converted into a vector with the same embedding model used to encode the Caregiving Knowledge Base. The DPR uses cosine similarity as distance function to calculate related documents in the vector space by calculating the user's question vector and existing document vectors inside the Chroma vector database. The most related documents were limited to three to keep the content relevant to the user's question, to avoid FM tokens window limitation, and to prevent degrading model generation quality [43].

Implementing CaLM using a RAG framework has two main benefits: It ensures that the model has access to the most current, reliable facts about caregiving, and it provides users with access to the model's sources, ensuring that its claims can be checked for accuracy and can ultimately be trusted. The caregiving knowledge base is designed to be updated regularly. The knowledge base in CaLM can be updated regularly and the model can be retrained more frequently. Updating the knowledge base and fine-tuning FM require fewer resources than retraining FMs. The model had access to the most current and reliable facts because of the frequency of the updates and the fine-tuning.

## Fine-Tuning FM in CaLM

CaLM uses a fine-tuned FM to synthesize an accurate answer tailored to the language and nuances of caregiving. The fine-tuned FM is an original FM that has been re-trained using supervised learning on Q-A pairs related to caregiving. Fine-tuning involves re-training the FM on a caregiving domain-specific dataset. Full fine-tuning that involves updating all of an FM's parameters is less feasible because it requires large computing resources. A more practical and commonly used type of fine-tuning is called Parameter-Efficient Fine-Tuning (PEFT), which requires retraining only part of or an extra component of the pretrained FM. The most widely used PEFT technique is called LoRA (Low-Rank Adaptation) [44], which adds a small number of trainable parameters to the FM while the original model parameters remain frozen. In developing CaLM, we used LoRA as well as the quantized technique of LoRA called QLoRA [45] to improve the FM's memory efficiency during re-training.

The main benefit of using RAG in CaLM is that by grounding an FM on a set of external, verifiable facts about caregiving, the model has fewer opportunities to pull information baked into its parameters, thus reducing the chances that the model will "hallucinate" incorrect or misleading information. This is critical in achieving the goal of a reliable CaLM.

## Data Sets for Fine Tuning FM and for Model Evaluation

Recent works indicate that an FM can be made more accurate by fine-tuning it on a high-quality smaller data set [46-49]. In this study, we used a large data set for RAG framework's knowledge base and a high-quality small data set for fine-tuning FM. The goal was to further improve the performance of the FM by retraining it using a high-quality data set tailored for caregiving Q-A so that it can respond more accurately and in a more contextually relevant way to questions related to caregiving. Fine-tuning is important particularly if we want to use a smaller and more efficient FM to achieve the goal of accessible CaLM.

The high-quality training set to fine-tune the FM was in the form of question-answer (Q-A) pairs, which were developed using a combination of both automatic and manual methods. Fragments of documents in the caregiving knowledge used in the RAG approach were randomly sampled and selected, and they were used as seed context to generate questions and answers. An OpenAI GPT-4 was used to generate variations of questions based on the seed contexts. This seed of questions was then paired with the three documents in the knowledge base that were most closely related to the questions to synthesize the output answer. The pairs of Q-A datasets were subsequently curated manually by a data annotator to validate that the questions were representative from a caregiver perspective and that the answers and their associated references were correct. The data annotator was responsible for selecting which Q-A pairs were relevant to caregiving. Following the curation, the dataset was deduplicated to remove duplicate data, to prevent data leakage, and to improve the fine-tuning process [49].

We constructed 581 Q-A pairs of datasets, each of which included questions and answers with references. Of these 581 pairs, we randomly selected 415 for inclusion in a training subset, reserving the remaining 66 pairs as a test subset (Table 3). The 415 Q-A pairs in the caregiving training set were used for fine tuning the FMs using supervised learning. The goal of the fine-tuning process is to adapt the pre-trained FM to the caregiving field so that it can respond more accurately and with more contextual relevance to questions related to caregiving. Once all the components of CaLM were developed and trained, the model was evaluated using the 66 pairs of questions that were not included in the training.

Table 3. Data sets for retraining during fine-tuning and for testing

| **Type** | **Subset** | **Examples** |
|---|---|---|
| Q-A With Inline Citation & References | Train Set | 415 |
| | Test Set | 66 |
| | Total | 581 |

We evaluated the performances of the models using the test set consisting of 66 Q-A pairs. The evaluation was conducted after the components of the RAG framework were developed and the small FMs were fine-tuned by retraining the models on the training set of 415 Q-A pairs. The training set (comprising 415 entries) and the testing set (comprising 66 entries) were distinct with no overlapping questions or content. The test set (comprising 66 entries) was distinct with no overlapping questions or content with the training set. After exposing the FMs to the questions from the test set, we checked the FMs' answers against the test set answers to determine whether the FMs were outputting accurate answers to the test questions. The FMs' performance was measured by the similarities between the reference answers to the test set questions and the output generated by the models. In addition to similarities, we evaluated the capabilities of the models in providing references to the answers.

We tested the small FMs trained in three different settings: Vanilla, RAG, and RAG + Fine-tuned. Vanilla is the original FM without adding the Caregiving Knowledge Base. In the RAG setting, for every question in the data, additional context from the knowledge base was added before the Q-A pair was sent to the FM. In the RAG + Fine-tuned setting, we followed the same procedure as with the RAG setting, except that the FM had already been retrained using the 415 Q-A training set pairs. We compared the small FMs with OpenAI GPT 3.5 as a baseline benchmark. Because GPT 3.5 is a proprietary commercial system, we were not able to control its fine-tuning variables other than providing examples. Therefore, we cannot compare it with other FMs that were retrained using the caregiving Q-A training set.

## Results

We applied the benchmark metrics commonly used in evaluating language models, including BLEU, ROUGE, CHAR-F, and BERT-score evaluation. These metrics are automatic evaluations that measure the similarity of a response to a provided reference answer in the test set. The metrics are considered to have a high correlation with human judgments of quality. BLEU (Bilingual Evaluation Understudy) is the oldest and most popular metric, and it captures word level similarities between the answer and the reference. ROUGE (Recall-Oriented Understudy for Gisting Evaluation) evaluates on how well the models produce generated answer compared to the reference answer by measuring several overlapping units such as n-gram, word sequences and word pairs [50]. The score for the BLEU and ROUGE metrics ranges between 0 and 1, with 1 being a perfect score. CHR-F (Character N-gram F-score) measures similarities at the character level, and it is scored between 0 and 100, with 100 being a perfect score. BERT (Bidirectional Encoder Representations from Transformers) calculates the similarity between output and a reference using sentence representation and focuses on measuring semantic similarity. The BERT metric is scored between 0 and 1, with 1 being a perfect score.

The purpose of using multiple metrics was to provide a more comprehensive intrinsic automatic evaluation and to assess whether the performances are consistent across different standards of granularity, from character, to word, to meaning. The performance is summarized in the chart in Figure 2. In addition to the general performance of the language models in providing the right answers, we evaluated their reliability by measuring their capacity to provide references with the answers. The capability of the models in providing the references is summarized in Table 4.

The RAG framework improved the performance of all FMs used in this study across all measures. The performances improved significantly at the character (CHAR-F), word (BLEU and ROUGE), and semantic (BERT) levels. As expected, the large FM (the estimated 175 billion parameters of OpenAI GPT 3.5) performed better than small FMs (Falcon 7B and LLaMA 2 7B with 7 billion parameters) across all metrics. The larger FM has more parameters and is able to accommodate having more knowledge encoded into it.

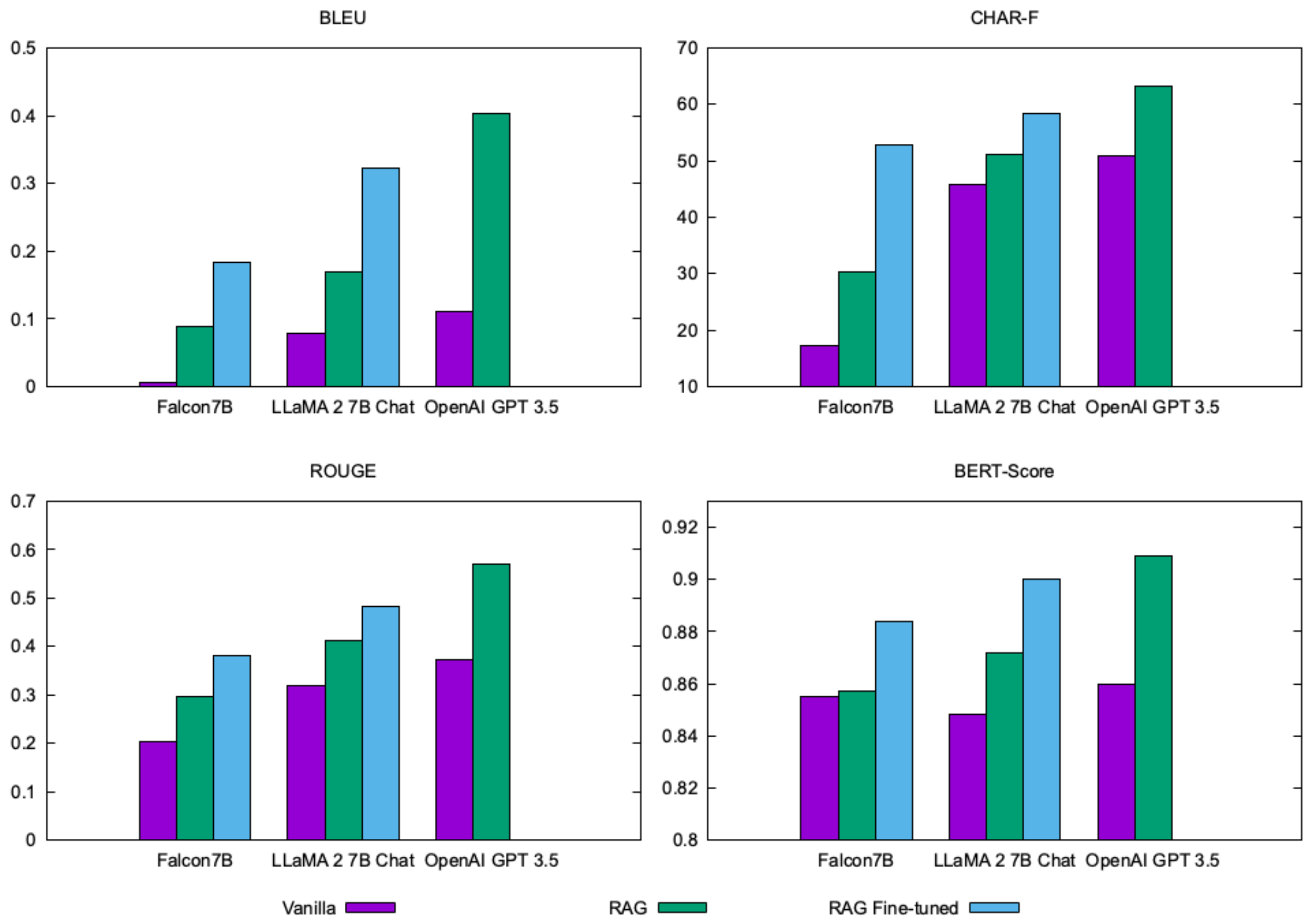

*Figure 2 Benchmarks of three models, Falcon7B, LLaMA2 7B Chat, and OpenAI GPT 3.5 on three different approach: Vanilla, with RAG, and RAG + Fine-tuned model (except GPT 3.5)*

The RAG framework that was implemented with fine-tuning performed better than RAG-only and Vanilla for the two small FMs. Because we could not retrain the GPT 3.5 model, we cannot evaluate a retrained (fine-tuned) version of it. The most interesting result is that LLaMA 2 7B RAG + Fine-tuned performed significantly better than Vanilla GPT 3.5 across all metrics. Fine-tuned Falcon 7B also performed better than Vanilla GPT 3.5 across all metrics. LLaMA 2 7B and Falcon 7B have only 7 billion parameters, while OpenAI GPT 3.5 has 175 billion parameters. This shows that a smaller FM with the injection of domain specific knowledge can perform better than a much larger FM.

In addition to the general performance of the language models in providing the right answers, we also evaluated their reliability by measuring their capacity to provide accurate references with the answers. Table 4 provides results on the capability of the models to return references in their answers, and the number of correct references. The references provided by the models were evaluated for correctness and relevance. In addition to checking whether the generated answer returned a list of references, the annotator verified that the links are correct and active. The annotator also checked each inline reference to determine if the content in the answer was part of the original document. Using the results from these checks, the data annotator decided whether each reference was correct and relevant to the answer.

None of the Vanilla FMs are trained to return references, and, therefore, no references were provided for any questions (the results were all zeroes). LLaMA 2 7B in the RAG + Fine-tuned setting provided references to all 66 answers. It performed better than GPT 3.5 with the RAG framework, which returned references on only 46 of the 66 answers (70%). Fine-tuning by retraining the FMs using the 415 Q-A pairs of the training set significantly improved the capabilities of the FMs to return references. The percentage of correct and relevant references among the models follow similar patterns with LLaMA 2 7B performing better than GPT 3.5 (80% vs. 62.1%). Table 2 of the provides examples of comparison between the generated answers provided by CaLM and GPT 3.5, as well as the reference answers from the Q-A test set.

Table 4. Capability in Returning Reference in the Answer

| **FM Model** | **Variants** | **Returning References** | **Not Returning References** | **Percentage of Returning References** | Percentage of Correct References (Human Evaluation) |
|---|---|---|---|---|---|
| Falcon 7B | RAG | 0 | 66 | 0% | 0% |
| | RAG Fine Tuned | 61 | 5 | 92% | 66% |
| LLaMA 7B - Chat | RAG | 17 | 49 | 26% | 12.1% |
| | RAG Fine Tuned | 66 | 0 | 100% | 80% |
| GPT 3.5 | RAG GPT 3.5 | 46 | 20 | 70% | 62.1% |

Implementing the RAG framework as we developed CaLM had several benefits. First, we were able to ensure that the language model had access to the most current facts. The caregiving knowledge could be updated regularly, and we could retrain the model more frequently than we would be able to using a new FM. Updating the knowledge base and fine-tuning FM require fewer resources than retraining FMs. Second, the model had access to the most reliable facts because of the frequency of the updates and the fine-tuning that included references in the knowledge base and the training set. The third benefit is that users have access to the model's information sources, ensuring that its answers can be checked for accuracy and ultimately trusted.

**Developing a caregiver chatbot using CaLM**

We developed a prototype of a caregiving chatbot that uses CaLM as the engine. The purpose of the chatbot prototype is to show that the CaLM is accessible and can be deployed in a low-resource environment. The chatbot interface and example of the interaction are illustrated in Figure 3. We developed the user interface and interaction system that work with the CaLM engine. The model that powers the chatbot is small and requires approximately 4-6 GB of memory. The chatbot can be deployed on a laptop or a small server.

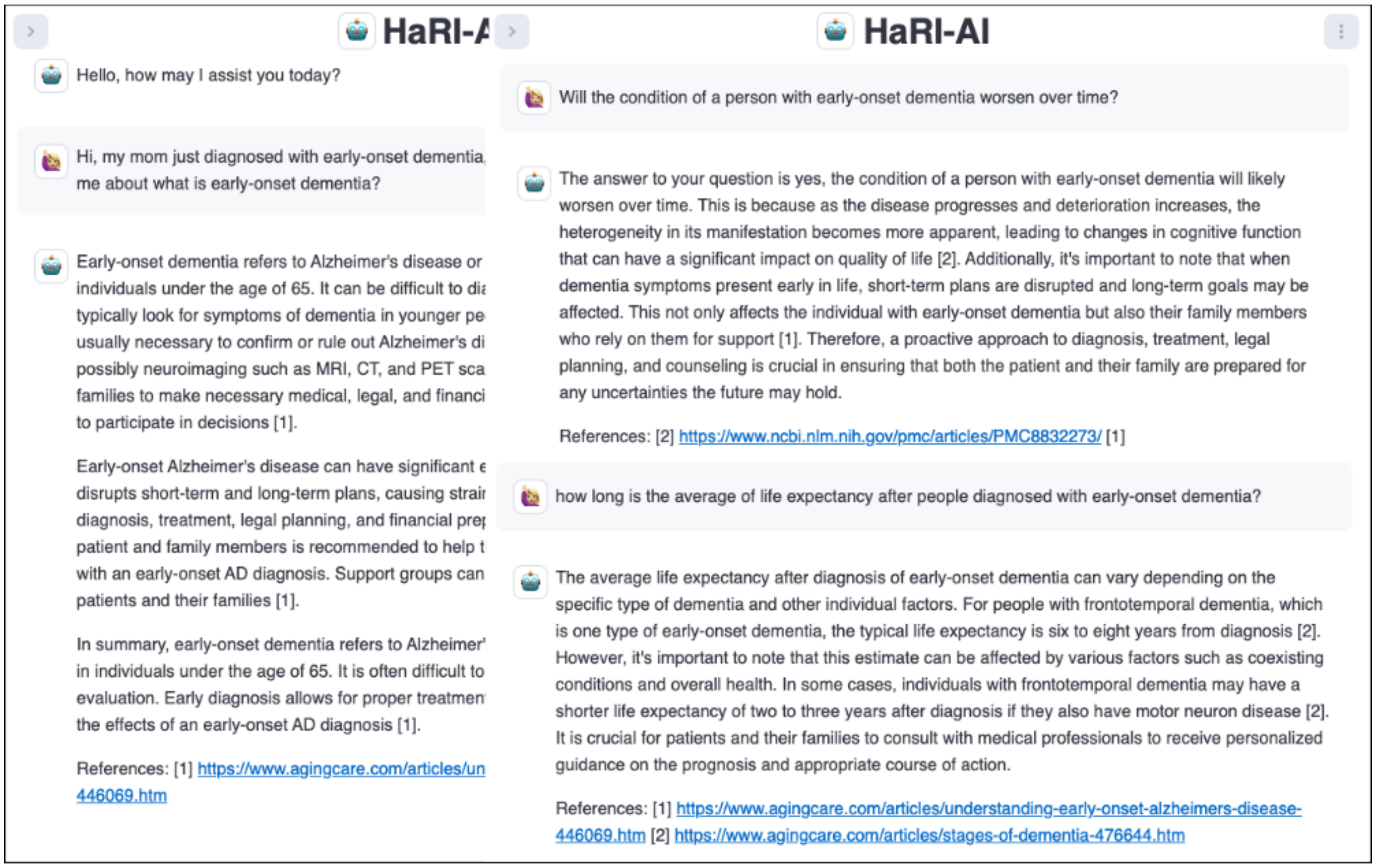

*Figure 3 Conversation example of CaLM Caregiving Chatbot*

# Discussion

## Principal Results

The findings of this study showed that a domain-specific language model for caregiving can be developed by using the RAG framework. The RAG framework improves the quality of answers generated by FMs in the domain-specific caregiving field by grounding the model on a caregiving knowledge base to supplement the general knowledge already stored in the internal representation of the FM. The results are consistent across different metrics, showing that FMs that are adapted using the RAG framework enriched with a caregiving knowledge base performed better than the original Vanilla FMs. Performance was further improved when the FMs were fine-tuned by retraining them using supervised learning on the specific Q-A training set related to caregiving. Fine-tuning also improved the reliability of the models by increasing their capability to provide verifiable responses though references. The result showed that a reliable CaLM can be developed by combining FM with RAG frameworks and fine-tuning that involves retraining the FMs.

The study found that small FMs can perform comparably to or better than much larger FMs when they are grounded in domain-specific knowledge related to caregiving. For example, LLaMA 2 7B (7 billion parameters) with access to a caregiver knowledge base performed better than GPT 3.5 with 175 billion parameters. Fine-tuned LLaMA 2 7B also provided reliable answers by supplying references to all of the answers in the test set. Smaller FMs require less computing power to train and fewer resources to deploy once they are trained. For example, LLaMA 7B with RAG and fine-tuning can be deployed using the computing power of small servers or desktop computers. This makes it more accessible for small organizations that want to develop language models specific to the domains important to them. Other methods for developing "small" FM that can perform almost as well as large FM have been proposed in recent works, one example is Starling-7B LLM [51] that uses Reinforcement Learning from AI Feedback (RLAIF) and high-quality large training data to enhance model performances.

The implications of the results show the potential for developing reliable and accessible language models in domain-specific areas by combining smaller fine-tuned FMs with the RAG framework, as evidenced by other studies [52, 53]. The parameter sizes available for FMs will continue to grow rapidly in the coming years. For example, the latest release of LLaMA 2 is available in three model sizes: 7, 13, and 70 billion parameters. The computing power accessible to smaller organizations will also continue to grow. Therefore, this approach will still have potential even when the size of FMs increases. By using an FM size that can be supported by the infrastructure available to small organizations, we can ensure that CaLM will remain accessible to those organizations. For example, FMs the size of 70 billion will likely be accessible within the next few years. Using smaller FMs with RAG + fine-tuning will remain a valid approach even as large FMs grow even larger because hallucinations will still exist in the large FMs.

Two main methods were used in the development of CaLM: RAG and fine-tuning of FM. Fine-tuning has a number of drawbacks, including the computational resources needed for full parameter fine tuning and the risk of catastrophic forgetting where the model lose its ability to perform well on the original task or domain after being fine-tuned with new data [54]. A previous study found that the PEFT technique is proven to be effective in preventing catastrophic forgetting on its original FM capabilities in comparison to the Full Parameters Fine-Tuning technique [55]. Therefore, fine-tuning with PEFT will increase model capability on CaLM's downstream tasks without sacrificing original FM capabilities.

Reliable and accessible CaLM has significant potential for solving downstream tasks through developing systems such as a chatbot for caregiving. This approach can also be used for developing systems for caregivers that are specific to the conditions of the care recipients they serve, such as for caregivers of individuals with disabilities or cancer. The contextual reference in the caregiving knowledge base can be further tailored to organizational needs or services. For example, if an organization has a service related to long-term care, the language model can be tailored using the organization's internal documents, procedures, and guidelines. We can also tailor the answer function to the desired communication style or education levels of the intended users.

## Limitations

While the study shows the potential of the development of a reliable and accessible language model in a specific domain such as caregiving, this study has a number of limitations. The first limitation is that the knowledge base is restricted to caregiving of individuals with ADRD. The choice to focus primarily on ADRD was driven by the significant global impact of this condition and the availability of extensive research and guidelines. However, this focus may limit the broader applicability of our findings to other caregiving contexts, potentially affecting the model's generalizability. To address this, future work will include expansion of the caregiving corpora related to other care recipient conditions, especially for individuals with chronic and complex conditions. The corpora related to issues and skills in general caregiving (irrespective of the care recipient's conditions) used in this study is also limited. We plan to expand the caregiving corpora by retrieving all publicly available documents related to caregiving. We envision CaLM as an iterative and evolving model. By integrating data and insights from broader caregiving contexts, we aim to evolve CaLM into a more inclusive and representative model, catering to the diverse needs and challenges encountered in caregiving.

The study used quantitative metrics for evaluating the performance of the language models. Quantitative metrics show consistency across different metrics. The benefits of quantitative metrics include reduced time and cost and increased consistency compared to human evaluation. All metrics used in this study fall into the category of intrinsic metrics [56]. Intrinsic metrics measure the proficiency of an LM in generating coherent and meaningful sentences relying on language rules and patterns [57]. However, these quantitative metrics are insufficient for capturing the multifaceted human perspectives and the practicalities encountered in real-world caregiving scenarios. Future research should include evaluation using extrinsic metrics that are crafted to encapsulate user experiences and the actual applicability of language models within real-world settings [57]. Extrinsic metrics that are relevant to the healthcare and caregiving domains should measure the accuracy and reliability of information, its timeliness and relevance, and the system's capability to provide empathetic and emotionally supportive responses [56]. The next phase of research will aim to incorporate these metrics and to engage real family caregivers and health care professionals to evaluate the quality of the answers. Because CaLM will be implemented as a chatbot that engages family caregivers, evaluation using psychological dimensions such as perceived humanness, likeability, anthropomorphism, animacy, and perceived safety [58] need to be included in more holistic evaluations in the future.

Additionally, ethical considerations are integral to the deployment of AI in caregiving. The AI system must employ stringent data privacy measures and transparency in its decision-making processes [59-61]. This transparency ensures that caregivers can trust and understand the rationale behind AI-generated advice and recommendations. Moreover, addressing potential biases in the training data is crucial to ensure that the AI system provides equitable support across all user demographics. Future development of CaLM will involve continuous engagement with stakeholders to address these ethical challenges and ensure the model's alignment with the highest standards of responsible AI practice.

## Conclusions

This study shows promise in the development of caregiving language models. It shows that CaLM developed using RAG frameworks and fine-tuning the FM can provide reliable answers to user questions by retrieving accurate references. The study shows that a reliable CaLM can be developed using FMs with a knowledge base specific to the caregiving domain. It also shows that a small FM that uses a caregiving knowledge base and is re-trained using caregiving Q-A performs better and more reliably than a much larger FM in answering caregiving-related questions. CaLM developed using small FMs performed better than the benchmark large FM (OpenAI GPT 3.5), which will allow CaLM to be accessible and deployed in low-resource settings. Future work includes expanding the domain knowledge to include other conditions of care recipients to enhance its utility. Furthermore, the evaluation process will be refined by engaging caregivers as end-users in providing feedback, alongside insights from healthcare professionals and caregiving domain experts.

## Funding Acknowledgement

This study was supported by grants from the National Institute on Disability, Independent Living, and Rehabilitation Research (NIDILRR grant number 90RTGE0002, *National Rehabilitation Research and Training Center on Family Support* and DPKT21000080 *Translating mHealth Technology to a Community Service Organization Providing Long Term Services and Supports*). NIDILRR is a Center within the Administration for Community Living (ACL), Department of Health and Human Services (HHS). The contents of this publication do not necessarily represent the policy of NIDILRR, ACL, or HHS, and readers should not assume endorsement by the US Federal Government.

**Contributions**

BP contributed to the conceptualization, methodology, data analysis, and writing of this study. BA contributed to the conceptualization, methodology, programming, data curation, data analysis, and writing. WS contributed to data curation, data cleansing, and Q-A annotation and validation. IMAS contributed to data analysis and machine learning infrastructure. YW and HH contributed to the data curation. AS contributed to the conceptualization and machine learning infrastructure. YC contributed to the conceptualization, methodology, writing, and editing.

**Conflict of Interest**

The authors declare no potential conflicts of interest with respect to the research, authorship, and/or publication of this article.

**Abbreviations**

ADRD: Alzheimer's Disease Related Dementias

BERT: Bidirectional Encoder Representations from Transformers

BLEU: Bilingual Evaluation Understudy

CaLM: Caregiver Language Model

FM: Foundation Model

GAI: Generative Artificial Intelligence

GPT: Generative Pre-trained Transformer

GPT: Generative Pre-trained Transformer

LLaMA: Large Language Model Meta AI

LLM: large language model

LoRA: Low-Rank Adaptation

PEFT: Parameter-Efficient Fine-Tuning

Q-A: Question-Answer

RAG: Retrieval Augmented Generation

ROUGE: Recall-Oriented Understudy for Gisting Evaluation

# Appendix

Table 1. Example of Q-A for Training (Fine-Tuning) FMs

| System Prompt | Question | Context | Answer |
|---|---|---|---|
| Generate a comprehensive, informative and helpful answer for given question solely based on the information provided (Source and Content). You must only use information from the provided contents. Combine content information together into a coherent answer. Do not repeat text. Anytime you cite piece of information for the answer, follow by add its numerical source in ```[]``` format after text you cite. Only cite the most relevant content that answer the question accurately. Format list of References in this markdown format ```[index](Source link)``` and added at the end. If a question does not make any sense, or is not factually coherent, explain why instead of answering something not correct. If provided information is not providing enough related information to answer the question don't make up an answer. If you | What communication strategies can I use to help my loved one with dementia understand and express their needs? | Content: How to communicate with a person with dementia? We aren't born knowing how to communicate with a person with dementia—but we can learn. Improving your communication skills will help make caregiving less stressful and will likely improve the quality of your relationship with your loved one. Good communication skills will also enhance your ability to handle the difficult behavior you may encounter as you care for a person with a dementing illness. Set a positive mood for interaction. Your attitude and body language communicate your feelings and thoughts more strongly than your words do. Set a positive mood by speaking to your loved one in a pleasant and respectful manner. Use facial expressions, tone of voice, and physical touch to help convey your message and show your feelings of affection. Get the person's attention. Limit distractions and noise—turn off the radio or TV, close the curtains or shut the door, or move to quieter surroundings. Before speaking, make sure you have her attention; address her by name, identify yourself by name and relation, and use nonverbal cues and touch to help keep her focused. If she is seated, get down to her level and maintain eye contact.Source: [1] <https://www.caregiver.org/resource/caregivers-guide-understanding-dementia-behaviors/>Content: What are some tips to communicate with a dementia patient? We aren't born knowing how to communicate with a person with dementia—but we can learn. Improving your communication skills will help make caregiving less stressful and will likely improve the quality of your relationship with your loved one. Good communication skills will also enhance your ability to handle the difficult behavior you may encounter as you care for a person with a dementing illness. Set a positive mood for interaction. Your attitude and body language communicate your feelings and thoughts more strongly than your words do. Set a | When communicating with a person with dementia, there are several strategies that can help them understand and express their needs. First, it is important to set a positive mood for interaction. This can be done by speaking to your loved one in a pleasant and respectful manner, using facial expressions, tone of voice, and physical touch to convey your message and show your feelings of affection [1][2]. Getting the person's attention is also crucial. Limit distractions and noise, such as turning off the radio or TV, closing the curtains or shutting the door, or moving to quieter surroundings. Before speaking, make sure you have their attention by addressing them by name, identifying yourself by name and relation, and using nonverbal cues and touch to help keep them focused. If they are seated, get down to their level and maintain eye contact [1]. When engaging in conversation, it is important to speak clearly and simply. Using hand gestures can also be helpful in conveying your message. It is best to talk about one thing at a time and be patient, understanding that some days may be better or worse than others [3]. In addition to communication strategies, it is important to manage sleep and behaviors in individuals with dementia. Sleep issues, wandering, confusion, and angry outbursts are common behavioral changes associated with dementia. To help manage these behaviors, it is recommended to secure all doors in the house to minimize wandering. Installing a surveillance system or alarms can also help keep your loved one safe. For sleep issues, it may be helpful to establish a consistent bedtime routine and create a calm and comfortable sleep environment [3]. Overall, improving communication skills and implementing |
| don't know the answer to a question, say you don't have answer and please don't share false information. | | positive mood by speaking to your loved one in a pleasant and respectful manner. Use facial expressions, tone of voice, and physical touch to help convey your message and show your feelings of affection.Source: [2] <https://www.caregiver.org/resource/ten-tips-for-communicating-with-a-person-with-dementia/>Content: or finding a local in-person caregiver support group. Doing so will put you in touch with other caregivers who can share their own experiences, listen to your frustrations, and offer their sage advice and answers. Improve communication Conversations with someone who has dementia can often feel challenging. Improve communication with your loved one by following these tips for talking with dementia patients : Limit distractions Speak clearly and simply; using hand gestures can help Maintain eye contact Talk about one thing at a time Be patient, and understand that some days may be better or worse than others Manage sleep and behaviors With dementia comes a myriad of behavioral changes, like sleep issues, wandering, confusion, and angry outbursts. These can all present major challenges for patients and caregivers alike. To help manage a loved one’s problematic behaviors, try the following tactics: To minimize wandering, make sure all doors in the house are secured, and consider installing a surveillance system or alarms to help keep your loved one safe. For sleep issuesSource: [3] <https://www.agingcare.com/articles/alzheimers-presents-unique-challenges-for-home-care-154829.htm> | strategies to manage behaviors can greatly enhance the caregiving experience and improve the quality of the relationship with your loved one with dementia. References:[1] https://www.caregiver.org/resource/caregivers-guide-understanding-dementia-behaviors/[2] https://www.caregiver.org/resource/ten-tips-for-communicating-with-a-person-with-dementia/[3] https://www.agingcare.com/articles/alzheimers-presents-unique-challenges-for-home-care-154829.htm |

Table 2. Example of CaLM and GPT 3.5 Generation Answer

| No | Question | Reference Answer | CaLM Answer (RAG + Fine-Tuned Model) | GPT3.5 Answer |
|---|---|---|---|---|
| 1 | What are the benefits of hospice care for individuals in the advanced stages of Alzheimer's disease? | Hospice care offers several benefits for individuals in the advanced stages of Alzheimer's disease. Firstly, hospice services are designed to support individuals at the end of life, providing specialized care and support during this challenging time. Hospice services may include support groups, visiting nurses, pain management, and home care. These services aim to improve the quality of life for the patient, ensuring their comfort and well-being.<br><br>In the late stage of Alzheimer's disease, individuals can be frail and susceptible to infections. Hospice care focuses on managing pain and other symptoms, helping the patient remain comfortable. This is particularly important as the side effects of many drug therapies can outweigh the benefits in this stage. Hospice teams are experienced in providing symptom management and can offer helpful suggestions to family and friends on how to connect with their loved one during the late stages of dementia.<br><br>Additionally, hospice care benefits families by offering caregiving assistance and emotional support. Caring for a loved one with advanced Alzheimer's disease can be overwhelming, and hospice services can provide much-needed support to the entire family. Hospice teams can help alleviate the burden of caregiving and provide emotional guidance during this difficult time.<br><br>In summary, hospice care provides specialized support and services for individuals in the advanced stages of Alzheimer's disease. It focuses on improving the patient's quality of life, managing symptoms, and providing assistance and emotional support to the family. | Hospice care can provide numerous benefits for individuals in the advanced stages of Alzheimer's disease, including support and comfort during the end-of-life process. It is important to note that hospice services are usually arranged through the treating physician and are typically not available until it is anticipated that a person has less than six months to live [1].<br><br>In terms of benefits for individuals in advanced stages of Alzheimer's disease, hospice care can help manage pain and other symptoms, providing relief from discomfort. Additionally, hospice teams can offer helpful suggestions on how family members and friends can continue to connect with their loved ones during the late stages of dementia [2].<br><br>Hospice services also provide support for families dealing with end-of-life care challenges. This includes offering emotional support and assistance in caring for individuals at home or in a skilled nursing facility/dementia care facility, where adequate management and supervision can be provided [1].<br><br>In summary, hospice care offers benefits such as pain management, symptom control, connection opportunities with loved ones during the late stages of dementia, emotional support for families, and assistance in caring for individuals at home or in a skilled nursing facility/dementia care facility. These advantages can greatly improve the quality of life for individuals in advanced stages of Alzheimer's disease.<br><br>References:<br>[1] https://www.caregiver.org/resource/alzheimers-disease-caregiving/<br>[2] https://www.agingcare.com/articles/can-dementia-be-fatal-476368.htm | Hospice care can provide numerous benefits for individuals in the advanced stages of Alzheimer's disease. Some of the key benefits include:<br>1. Comfort and pain management: Hospice care focuses on making the individual as comfortable as possible, managing their pain and ensuring their quality of life. This is particularly important for individuals with Alzheimer's disease who may experience discomfort or pain due to physical symptoms or complications.<br>2. Emotional and psychological support: Hospice care teams include professionals such as social workers, counselors, and chaplains who provide emotional and psychological support not just to the patient but also to their family members. They help individuals cope with the challenges and emotional stress that come with Alzheimer's disease.<br>3. Customized care plans: Hospice care teams work closely with the patient, family, and medical professionals to develop a personalized care plan that meets the specific needs and preferences of the individual. They consider factors such as the stage of Alzheimer's disease, cognitive abilities, and any other existing medical conditions.<br>4. Enhanced quality of life: By providing specialized care, hospice teams aim to improve the quality of life for individuals in the advanced stages of Alzheimer's disease. They focus on maintaining dignity, maximizing comfort, and promoting a sense of well-being for the patient.<br>5. Assistance with daily tasks: As Alzheimer's disease progresses, individuals may struggle with performing essential activities of daily living (ADLs) such as eating, bathing, and dressing. Hospice care can provide assistance with these tasks, helping individuals maintain their independence and dignity for as long as possible.<br>6. Family education and guidance: Hospice care teams educate and guide family members on how to best care for their loved ones with Alzheimer's disease. They provide information on symptom management, communication techniques, and strategies for providing comfort and support.<br>7. Bereavement support: When the individual with Alzheimer's disease eventually passes away, hospice care teams continue to provide support to the family during the grieving process. They offer counseling, support groups, and resources to help family members navigate their loss and find healing.<br>It's important to note that hospice care is typically provided in the individual's home or in a specialized hospice facility, allowing them to receive care in a familiar and comfortable environment. The goal of hospice care is to improve the overall well-being of individuals with advanced Alzheimer's disease and support their families throughout the progression of the disease. |
| 2 | What are the early signs of dementia? | The early signs of dementia can vary from person to person, but there are common warning signs to watch for. Memory loss is a key sign of dementia, where individuals may repeat themselves or rely on aids such as sticky notes to remember things. Forgetting the names of familiar friends and family members is also a sign of dementia. It's important to note that memory loss from dementia is not a normal part of aging, and forgetting entire events due to short-term memory loss is cause for concern. | The early signs of dementia can vary depending on the cause, but there are some common symptoms to watch for. Alzheimer's disease is the most common type of dementia and its warning signs may be familiar. However, symptoms can go unnoticed for a while even if you are caring for someone with Alzheimer's [1].<br><br>Some early signs of Alzheimer's include wandering and getting lost, repeating questions or sentences, difficulty recognizing friends and family, impulsive behavior, and experiencing hallucinations or delusions. It is important to note that these symptoms may be similar to other types of dementia [1].<br><br>In addition to Alzheimer's disease, there are other types of dementia with distinct early warning signs. Lewy body dementia has some symptoms similar to Alzheimer's but also includes experiencing hallucinations or delusions, disorganized or illogical ideas, muscle rigidity, and reduced facial expression [1].<br><br>Frontotemporal dementia is another type of dementia with different early warning signs. It may include experiencing hallucinations or delusions, disorganized or illogical ideas, muscle rigidity, and insomnia [1].<br><br>In summary, the early signs of dementia can vary depending on the cause but common symptoms to watch for include memory loss, repeating questions or sentences, difficulty recognizing friends and family, impulsive behavior, experiencing hallucinations or delusions, disorganized ideas, muscle rigidity, reduced facial expression, and insomnia. If you notice any of these signs in your loved one, it is important to consult with a doctor who can run tests to determine the cause [1].<br><br>References:<br>[1] https://www.agingcare.com/articles/alzheimers-disease-dementia-warning-signs-144253.htm | Dementia is a progressive neurological disorder that affects cognitive function, memory, and behavior. While the symptoms of dementia may vary depending on the type and stage of the disease, there are some early warning signs that may indicate the presence of dementia. These signs can include:<br><br>1. Memory loss: Difficulty remembering recent events or conversations, forgetting important dates or appointments, and relying on memory aids or family members for reminders.<br>2. Difficulty with language and communication: Struggling to find the right words, repeating phrases or stories, or having trouble following and participating in conversations.<br>3. Confusion and disorientation: Becoming disoriented in familiar surroundings, getting lost on familiar routes, or having trouble understanding time, date, or location.<br>4. Impaired judgment and decision-making: Exhibiting poor judgment or making uncharacteristic decisions, such as giving away money or falling for scams.<br>5. Decline in problem-solving abilities: Having difficulty with tasks that require planning, problem-solving, and critical thinking skills, such as managing finances or following a recipe.<br>6. Personality and mood changes: Experiencing changes in mood, behavior, or personality, such as becoming more agitated, irritable, anxious, or apathetic.<br>7. Withdrawal from activities and social interactions: Losing interest in hobbies or social activities, avoiding social interactions, or becoming less engaged in previously enjoyed activities.<br><br>It's important to note that these signs can also be caused by other factors and may not necessarily indicate dementia. If you or someone you know is experiencing these symptoms, it is recommended to consult with a healthcare professional for a proper evaluation and diagnosis. |
| 3 | How can I ensure my loved one's safety at home, especially if they have a tendency to wander or become agitated | If your loved one has a tendency to wander or become agitated, there are several steps you can take to ensure their safety at home. First, it is important to reassure the person and speak to them calmly, listening to their concerns and frustrations. Allowing the person to maintain as much control in their life as possible is also important. Respecting their personal space and providing quiet times throughout the day, along with engaging activities, can help create a sense of security. Keeping well-loved objects and photographs around the house can also help the person feel more secure.<br><br>In terms of addressing wandering behavior, it may be helpful to identify triggers in the person's | To ensure the safety of your loved one at home, particularly when they tend to wander or become agitated, there are several steps you can take.<br><br>1. Reassure and speak calmly: Speak in a soothing voice and address their concerns and frustrations. Listen actively without interrupting them [2].<br>2. Allow control over personal space: Respect the person's autonomy by allowing them to maintain their personal space as much as possible [1].<br>3. Establish quiet times and engaging activities: Incorporate periods of calmness throughout the day, along with stimulating activities that they enjoy. This can help prevent restlessness and wandering [2]. | Ensuring the safety of a loved one who has a tendency to wander or become agitated can be challenging, but there are several steps you can take to create a safer environment for them:<br><br>1. Secure the home: Install secure locks on doors and windows to prevent your loved one from leaving without your knowledge. Consider using alarms or chimes that will alert you if a door or window is opened. You may also want to consider adding deadbolts to exterior doors.<br>2. Remove hazards: Identify and remove any potential hazards in the home that could cause harm to your loved one. This may include removing loose rugs or furniture, securing electrical cords, and keeping the floors clear of clutter. |
| | | environment and make changes accordingly. For example, if the person tends to wander more in the evening hours, it could be attributed to activities they used to be responsible for during that time of day, such as getting the kids from school or fixing dinner. Redirecting the person with another activity before they start wandering can be effective. Engaging them in activities that provide a sense of purpose and accomplishment, such as folding laundry or sorting coins, can help alleviate restlessness. However, it is important to monitor all activities to ensure safety.<br><br>By implementing these strategies, you can help ensure the safety of your loved one at home, minimizing the risk of wandering and agitation. | 4. Maintain familiar objects and photographs: Keeping well-loved items around the house provides a sense of security for your loved one [1].<br>5. Remind without prompting memory loss: If necessary, gently remind them who you are without pressing their memory too much. Avoid making assumptions about what they may or may not remember [2].<br>6. Encourage two-way conversations: Engage in active listening and allow for a back-and-forth conversation as long as possible. This can help maintain their sense of connection with you [1].<br>7. Distract through familiar activities: If communication becomes challenging, try distracting them with an activity they enjoy or find comfort in, such as reading a book or looking at photographs [2].<br><br>By implementing these strategies, you can create a safer environment for your loved one at home while also promoting their emotional well-being. However, if wandering becomes a significant concern and cannot be managed with the above measures, it may be necessary to explore additional safety solutions such as installing locks or alarms on doors that they tend to wander towards [2].<br><br>References:<br>[1] Tips for Caregivers and Families of People With Dementia (alzheimers.gov)<br>[2] https://www.agingcare.com/articles/wandering-top-tips-how-to-minimize-agitation-and-restlessness-432424.htm | 3. Utilize technology: Consider using technology to aid in their safety. For instance, you could install a video monitoring system to keep an eye on them remotely. There are also wearable devices available that can track their location or send alerts if they wander.<br><br>4. Establish a routine: Creating a predictable routine can help reduce agitation and confusion. Stick to a schedule for meals, medication, and activities. Having a consistent routine can provide a sense of comfort and stability.<br>5. Provide visual cues: Use visual cues throughout the home to help orient your loved one and provide reminders of important information. For example, you can place signs on doors to indicate the bathroom or bedroom, or use labels on drawers and cabinets to assist with finding items.<br>6. Ensure proper lighting: Good lighting is crucial to prevent falls and disorientation. Make sure all areas of the home are well lit, especially hallways, staircases, and entryways. Consider using night lights in the bedroom and bathroom to reduce confusion during nighttime.<br>7. Consider professional help: If your loved one's wandering or agitation becomes a constant concern, it might be beneficial to seek professional help. Consult with their healthcare provider or a geriatric specialist who can provide guidance and recommend appropriate interventions.<br><br>Remember, each situation is unique, and it's essential to tailor the safety measures to your loved one's specific needs. Regularly reassess their safety needs and make adjustments accordingly. |
| 4 | How do I obtain legal guardianship or power of attorney for a loved one with Alzheimer's? | To obtain legal guardianship or power of attorney for a loved one with Alzheimer's, it is crucial to make legal preparations for who will handle their medical and financial decisions on their behalf. Two important legal documents to consider are a durable power of attorney (POA) for health care and a durable power of attorney for finances. These documents appoint a trusted individual to oversee these areas of the patient's life once they are incapacitated.<br><br>To obtain these legal documents, it is recommended to consult with an elder law attorney who can assist in drawing up the POA documents and other essential estate planning documents such as wills, advance directives, and trusts. An elder law attorney specializes in legal matters related to the elderly and can provide guidance and expertise in navigating the legal process.<br><br>It is important to make these legal preparations as soon as possible because once a dementia patient is declared legally incompetent, they are unable to | To obtain legal guardianship or power of attorney for a loved one with Alzheimer's, you can follow these steps:<br><br>1. Consult an Elder Law Attorney: It is recommended to consult an elder law attorney who specializes in estate planning and dementia-related matters [1]. They can guide you through the legal process of obtaining power of attorney for health care or finances, which are critical documents that appoint a trusted individual to handle medical and financial decisions on behalf of your loved one once they become incapacitated.<br>2. Determine Which Type of Power of Attorney: There are two types of durable powers of attorney commonly used for individuals with Alzheimer's disease - power of attorney (POA) for health care and POA for finances [1]. The first type focuses on making medical decisions, while the second pertains to financial matters. It is important to determine which one you need based on your loved one's specific situation.<br>3. Gather Required Documentation: Depending on the state or jurisdiction where your loved one resides, there may be specific requirements for obtaining power of attorney [1]. Your elder law attorney can inform you about these | Obtaining legal guardianship or power of attorney for a loved one with Alzheimer's can be a complex process, but it is essential to ensure their well-being and protect their interests. Here are the general steps to take:<br><br>1. Research local laws: Familiarize yourself with the laws and procedures regarding guardianship and power of attorney in the jurisdiction where your loved one resides. Laws can vary, so it's important to understand the specific requirements and processes.<br>2. Consult an attorney: Seek legal advice from an attorney experienced in elder law or estate planning. They can guide you through the legal steps and help you understand the implications and responsibilities of becoming a guardian or obtaining power of attorney.<br>3. Determine capacity: Determine if your loved one has the legal capacity to make decisions. If they are still capable of understanding and making informed choices, it may be possible to establish a power of attorney. However, if their cognitive abilities are significantly impaired, guardianship might be necessary.<br>4. File a petition: If guardianship is required, you'll likely need to file a petition with the appropriate court. The court will evaluate the individual's capacity and determine if |
| | | do so themselves. Taking proactive steps to establish legal guardianship or power of attorney can help ensure that the necessary decisions and arrangements are made in the best interest of the loved one with Alzheimer's.<br><br>In addition to legal preparations, it is also important to encourage Alzheimer's patients to stay mentally and physically active, as these factors may help slow memory loss and delay declines in functional abilities. Engaging in activities that the patient enjoys on a regular basis can be beneficial. Adopting a healthy diet is also recommended for overall well-being. | particular needs and help ensure that all necessary documents are prepared correctly.<br>4. Appoint a Trustworthy Individual: Once you have obtained the appropriate type of power of attorney, it is crucial to appoint an individual who is trustworthy and capable of handling your loved one's medical or financial matters [1]. This person should be someone who has their best interests at heart and can make informed decisions on their behalf.<br>5. Consider Estate Planning Essentials: In addition to power of attorney documents, it may be beneficial to consult with an elder law attorney regarding other estate planning essentials such as wills, advance directives, and trusts [1]. These legal elements can help ensure that your loved one's wishes are respected and their assets are managed properly.<br>6. Act Urgently: It is important to note that once a dementia patient is declared legally incompetent, they cannot make decisions regarding medical or financial matters themselves [1]. Therefore, it is crucial to act urgently when seeking legal guardianship or power of attorney for your loved one with Alzheimer's disease.<br><br>By following these steps and consulting an experienced elder law attorney, you can obtain the necessary legal documents that will allow someone trustworthy to handle medical and financial decisions on behalf of your loved one with Alzheimer's disease.<br><br>References:<br>[1] https://www.agingcare.com/articles/what-happens-after-alzheimers-diagnosis-154289.htm | guardianship is necessary. This process may involve providing medical evaluations and other supporting documents.<br>5. Court evaluation: A court may appoint an independent evaluator, such as a physician or psychologist, to assess the individual's mental capacity and determine if guardianship is appropriate. The court will consider the evaluator's report when making a decision.<br>6. Attend a hearing: Once the petition is filed, you may need to attend a court hearing. During this hearing, you'll present your case, explaining why guardianship is necessary and why you should be appointed as the guardian. Other interested parties, such as relatives or friends, may also have the opportunity to voice their opinions.<br>7. Obtain court order: If the court determines that guardianship is necessary and that you are a suitable guardian, they will issue a court order granting you legal authority to act on behalf of your loved one. This order outlines your responsibilities and the extent of your decision-making authority.<br><br>It's important to note that laws and procedures can vary by jurisdiction, so it's advisable to consult with a legal professional who can provide guidance specific to your situation. They can help ensure that you follow the necessary steps and fulfill the legal requirements to obtain guardianship or power of attorney for your loved one. |